\documentclass[letterpaper]{article} 
\usepackage[preprint]{aaai2027}  
\usepackage[hyphens]{url}  
\usepackage{graphicx} 
\usepackage{natbib}  
\usepackage{caption} 
\usepackage{algorithm}
\usepackage{algorithmic}

\usepackage{amsmath}

\usepackage{booktabs}

\usepackage{multirow}
\usepackage[table]{xcolor}

\newcommand{\ourmethod}{HAM-VLN}

\title{\ourmethod: Harnessing Hierarchical Agentic Memory for Zero-Shot \\ Vision-and-Language Navigation}

\author{
    An Liu\textsuperscript{\rm 1, 2},
    Bingxi Liu\textsuperscript{\rm 3, 5},
    Hongyu Ding\textsuperscript{\rm 1, 4},
    Yixuan Jiang\textsuperscript{\rm 1, 4},
    Yaran Chen\textsuperscript{\rm 6},
    Fulin Tang\textsuperscript{\rm 1, 2, 7}\textsuperscript{*},\\
    Cong Leng\textsuperscript{\rm 7},
    Hong Zhang\textsuperscript{\rm 5},
    Jian Cheng\textsuperscript{\rm 1, 2}
}
\affiliations{
    \textsuperscript{\rm 1}Institute of Automation, Chinese Academy of Sciences, Beijing, China\\
    \textsuperscript{\rm 2}School of Artificial Intelligence, University of Chinese Academy of Sciences, Beijing, China\\
    \textsuperscript{\rm 3}EngineAI, Shenzhen, China\\
    \textsuperscript{\rm 4}Nanjing University, Nanjing, China\\
    \textsuperscript{\rm 5}Southern University of Science and Technology, Shenzhen, China\\
    \textsuperscript{\rm 6}Xi'an Jiaotong-Liverpool University, Suzhou, China\\
    \textsuperscript{\rm 7}MAICRO, Nanjing, China\\
    \{liuan2026@ia, tangfulin2015@ia, jcheng@nlpr.ia\}.ac.cn, \{12231061, hzhang\}@sustech.edu.cn,

    \{hongyuding, yixuanjiang\}@smail.nju.edu.cn, Yaran.Chen@xjtlu.edu.cn, lengcong@airia.cn,
    \textcolor{blue}{\urlstyle{tt}\url{https://an1iu.github.io/HAM-VLN}}
}
\begin{document}

\maketitle

\begingroup
\renewcommand{\thefootnote}{*}
\footnotetext{Corresponding authors.}
\endgroup

\begin{abstract}

Vision-and-language navigation (VLN) enables robots to follow instructions in previously unseen environments. Recently, a training-free paradigm has emerged: the robot queries a multimodal LLM to understand its observations and plan the next action. However, long-horizon navigation based on either image streams or dense map inevitably introduces a growing memory and reasoning bottleneck. We present \ourmethod, a decision-coupled, agent-authored memory that equips the robot with a persistent, depth-grounded world graph. In the same model call used to select the next action, \ourmethod{} also records semantic and reflective information---including room type, objects, navigation progress, and failure notes. Recent waypoints remain verbatim within a bounded window, while older history re-enters the context only through retrieval scored by relevance, recency, and salience, together with one-hop topological expansion. This design requires no additional LLM calls beyond the per-waypoint decision. Compared to previous methods, \ourmethod{} not only improves various navigation metrics but also reduces the context length by more than 65\%. Specifically, \ourmethod{} achieves 61.0\% Success Rate (SR) on VLN-CE R2R, 52.7\% SR on VLN-CE RxR, and 79.7\% SR on HM3D-v2 ObjectNav without any training. 
\end{abstract}

\section{Introduction}
\label{sec:intro}

Vision-and-Language Navigation (VLN) asks a robot to follow a natural-language instruction through an unseen environment and stop at the intended location or object~\citep{anderson2018vision,krantz2020beyond,gu2022vlnsurvey}. Multimodal large language models (MLLMs) can provide the robot with agentic reasoning without any training: at each waypoint, the robot uses a model to understand its observation and choose the next action~\citep{zhou2024navgpt}. However, each model call can reason only over the information presented to it. Over a long trajectory, later decisions need to know where the robot has been, how the instruction has progressed, and which branches have failed. Providing this information without continually expanding the context is the memory bottleneck.

\begin{figure}[t]
\centering
\includegraphics[width=\columnwidth]{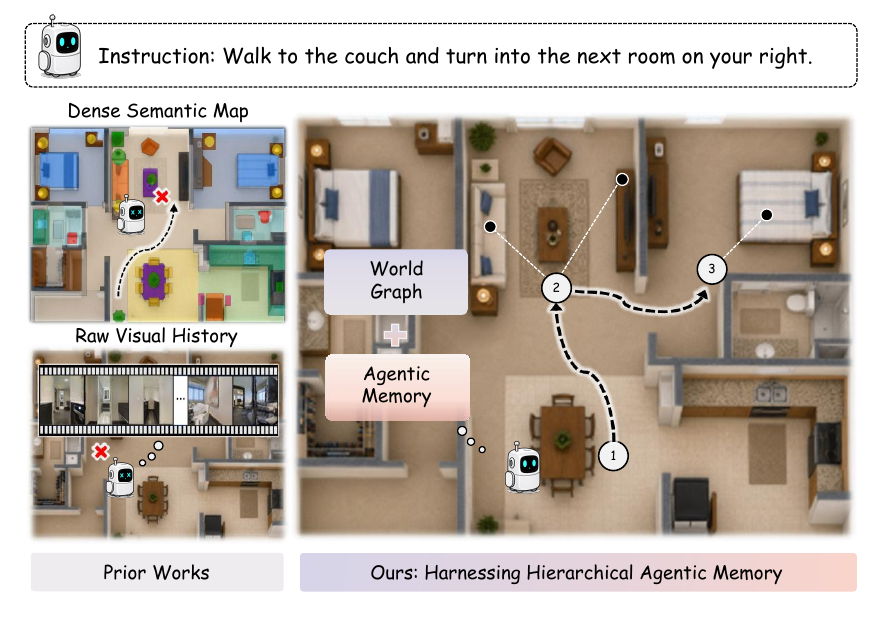}
\caption{Prior systems (left) present navigation history through either a \emph{dense semantic map}, whose storage grows with area, or the \emph{raw visual history}, whose context grows with the trajectory. \ourmethod\ (right) stores long-term navigation history in a depth-grounded world graph and uses hierarchical agentic memory to determine what remains in context and what is retrieved for each decision.}
\label{fig:teaser}
\end{figure}

There are two common representations of navigation history in zero-shot VLN. \textbf{Dense semantic maps} store tensors aligned to pixels or voxels~\citep{chaplot2020semexp}. Their size grows with the mapped area, and their contents---whether closed-category labels or open-vocabulary features~\citep{huang2023vlmaps}---are built separately from the navigation decision. Instructions such as ``the table in the bedroom'' or ``the hallway toward the kitchen'' must then be matched to metric cells. \textbf{Raw visual histories} instead append past panoramas to the model context. Important landmarks become buried in other visual details, context grows at every step, and a past dead end is not marked as a branch to avoid. Both make past observations available, but the MLLM does not decide what to retain for later decisions.

\ourmethod\ brings agentic memory into the robot's navigation loop. It keeps observations from the most recent $K$ waypoints in working memory and stores older experience in a depth-grounded world graph. When the MLLM selects the robot's high-level action, the same call records place semantics, observed objects, instruction progress, and failure evidence. For each subgoal, a local encoder ranks places by relevance, recency, and salience, and \ourmethod\ includes their topological neighbors. This design adapts the memory principles of generative agents~\citep{park2023generative} by linking memory to the robot's spatial experience and current subgoal. 

Our contributions are as follows:

\begin{enumerate}
\item \textbf{A world graph for robot navigation.} We represent navigation history as a graph of places and objects located using depth, with edges that record topology. The graph grows with the places and objects discovered by the robot rather than with the mapped area, while preserving spatial and semantic information for navigation.
\item \textbf{Hierarchical agentic memory.} The MLLM call selects the robot's action and records semantic state and failure notes. Working, episodic, semantic, and reflection views determine what remains in context, what is retrieved, and how failure evidence is reused, without extra LLM calls for memory maintenance.
\item \textbf{Results and memory efficiency.} Without any training, the same configuration reaches $61.0\%$ and $52.7\%$ success on R2R-CE and RxR-CE, respectively, and $79.7\%$ success on HM3D-v2 ObjectNav. On R2R-CE, \ourmethod\ improves all four navigation metrics over full raw history while reducing per-episode context by $67\%$.
\end{enumerate}

\begin{figure*}[t]
\centering
\includegraphics[width=0.85\textwidth]{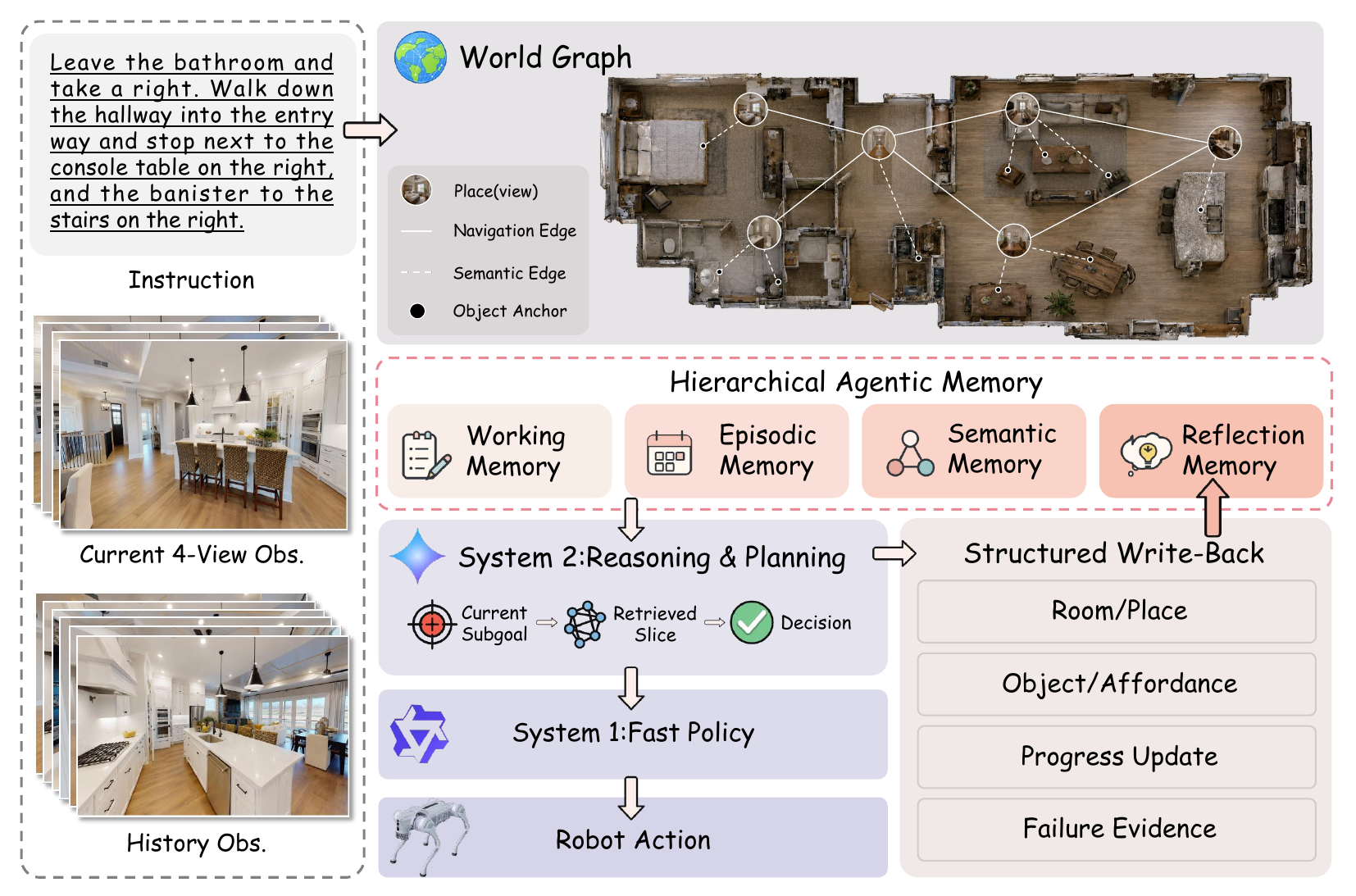}
\caption{\ourmethod\ overview. At each waypoint, System~2 reads the current panorama, the bounded $K$-waypoint working memory, and the retrieved graph slice, then emits the navigation action \emph{and} the memory writes (room type, objects, progress, failure notes) in one structured call; System~1 grounds the chosen direction into pixels, and the controller executes. The current subgoal retrieves a task-relevant slice that re-enters the next planning context.}
\label{fig:pipeline}
\end{figure*}

\section{Related Work}
\label{sec:related}

\paragraph{Vision-and-Language Navigation.}
Vision-and-Language Navigation (VLN) requires a robot to follow a natural-language instruction through a previously unseen environment~\citep{anderson2018vision,krantz2020beyond}. While trained navigators learn temporal representations from navigation trajectories~\citep{fang2019smt,zhang2024navid,wei2025streamvln}, recent zero-shot VLN methods use pretrained MLLMs to decompose instructions, reason over observations, refine plans, and select actions~\citep{zhou2024navgpt,long2024instructnav,chen2025constraint,ding2025lavira}. Zero-shot navigators must therefore expose trajectory state through observations, summaries, maps, or external memory. Over long trajectories, preserving visited places, instruction progress, and failed exploration without continually expanding the planning context becomes a central challenge.

\paragraph{Agentic Memory and Reflection.}
Long-horizon LLM agents address decision continuity by externalizing past experience into persistent memory. Generative agents organize a memory stream through relevance, recency, and importance, together with reflection over accumulated experience~\citep{park2023generative}; MemGPT manages bounded model context through external memory~\citep{packer2023memgpt}, while Reflexion retains verbal feedback from unsuccessful attempts to guide later behavior~\citep{shinn2023reflexion}. Embodied navigation adds a spatial requirement to this process. A useful memory must preserve where objects were observed, how visited regions connect, how the instruction has progressed, and where previous decisions failed.

\paragraph{Grounded Memory for Embodied Navigation.}
Navigation systems provide history to the planner through representations with different balances of perceptual detail, spatial structure, and context cost. Raw visual histories retain direct evidence but grow with the trajectory and offer no explicit index over places, task progress, or failed branches. Dense semantic maps provide persistent metric structure~\citep{chaplot2020semexp}, while open-vocabulary maps and online 3D scene graphs make objects and relations more accessible to language-based reasoning~\citep{huang2023vlmaps,yin2024sgnav}; 3D scene graphs have also served as substrates for LLM-based robot planning~\citep{rana2023sayplan}. Recent navigators further incorporate hierarchical memory and memory-augmented reasoning into zero-shot decision making~\citep{lyu2026himemvln,wu2026rememnav}. \ourmethod\ maintains semantic, progress, and failure state alongside navigation decisions, associates that state with a depth-grounded world graph, and recalls a subgoal-relevant graph slice for subsequent planning. Past experience returns as selected navigation state rather than as an accumulated visual transcript.

\begin{figure*}[t]
\centering
\includegraphics[width=0.85\textwidth]{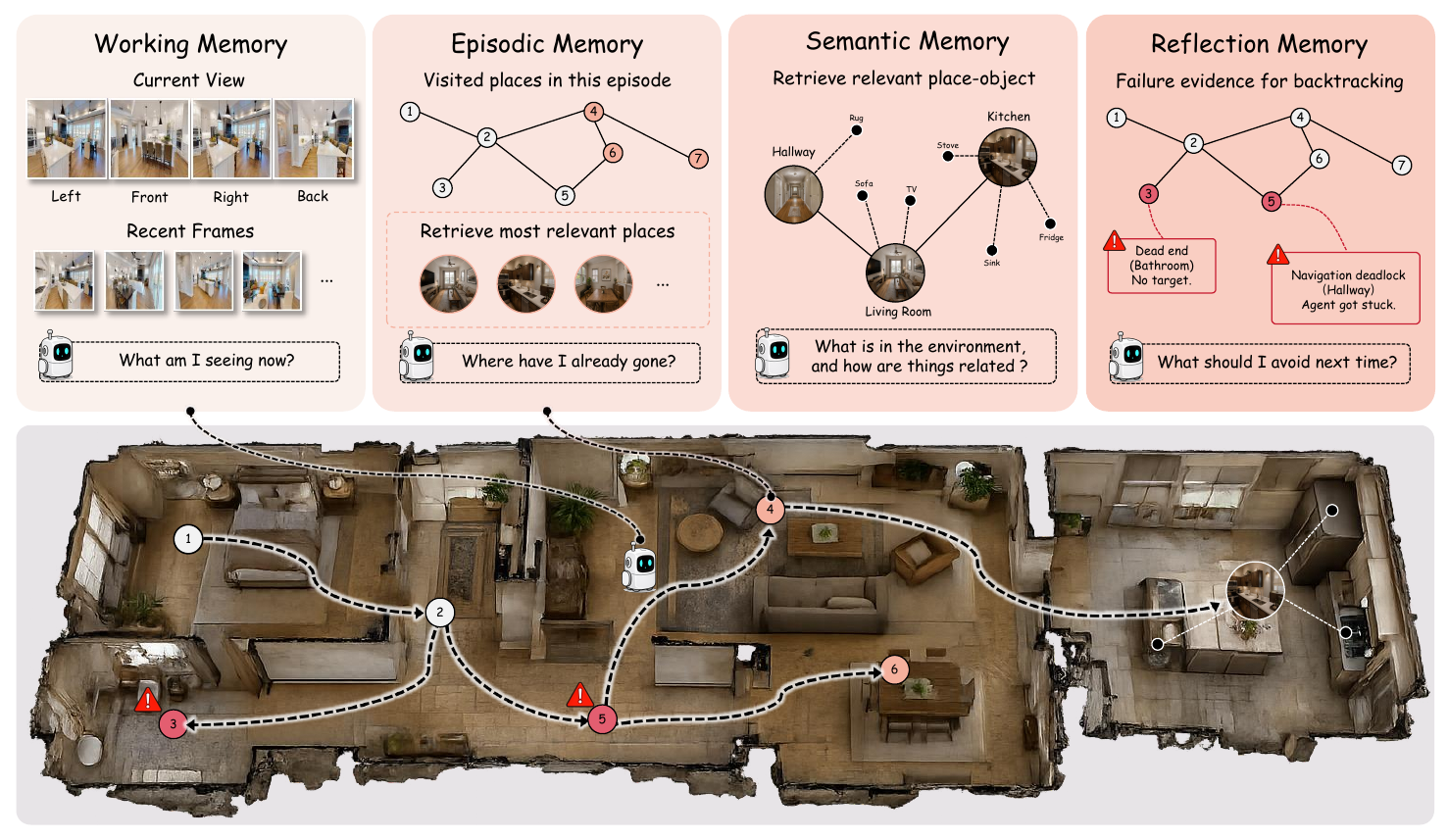}
\caption{Four functional memory views over one episode. (1)~\emph{Working memory}: the recent egocentric frame stream inside the $K$-waypoint window---what the planner sees verbatim. (2)~\emph{Episodic memory}: temporal retrieval over visited places, supplying recency. (3)~\emph{Semantic memory}: content retrieval over the place--object graph, supplying relevance. (4)~\emph{Reflection memory}: notes attached to abandoned places that resurface verbatim on retrieval.}
\label{fig:memory}
\end{figure*}

\section{Method}
\label{sec:method}

\ourmethod\ equips a zero-shot VLN agent with decision-coupled memory that preserves task-relevant state over long trajectories. Within its dual-process architecture, a world graph records agent-authored state and supports subgoal-conditioned retrieval and grounded backtracking, turning failed exploration into reusable evidence. Figure~\ref{fig:pipeline} illustrates how these components interact throughout a navigation episode.

\subsection{Dual-Process Architecture}
\label{sec:method-interface}

Deploying MLLMs for continuous navigation requires converting high-level semantic decisions into executable actions grounded in the robot's current visual and geometric state. Inspired by hierarchical VLA~\cite{shi2025hi} and dual-system VLN~\cite{wei2025ground}, \ourmethod~adopts a dual-process architecture. A slow, deliberative \emph{System~2} determines where to go and why, while a fast, reactive \emph{System~1} grounds this high-level intent in the current observation by predicting a pixel-space target, which is subsequently converted into executable motion commands. The policy at each step is factorized as
\begin{equation}
A_t = \pi_{\mathrm{robot}}\big(\varphi_{1}\big(\varphi_{2}(\mathcal{T}, O_t, M_t)\big),\, D_t, P_t, \mathbf{K}_{\mathrm{cam}}\big),
\label{eq:policy}
\end{equation}
where $A_t \in \mathcal{A}_{\mathrm{robot}}$ denotes the low-level action executed at step $t$ in the simulator's executable action space. Here, $\mathcal{T}$ denotes the navigation instruction, $O_t$ the observation at the current waypoint, $M_t$ the retrieved memory context, $D_t$ the corresponding depth observation, $P_t$ the robot's odometric pose, and $\mathbf{K}_{\mathrm{cam}}$ the camera intrinsic matrix. In the following, we use the lowercase $a_t$ to denote the high-level decision produced by System~2.

\paragraph{System~2: reasoning and planning.}
$\varphi_2$ is a large MLLM invoked once per waypoint---the pose where the robot stops, looks around, and decides. Reading $\mathcal{T}$, $O_t$, and $M_t$, it commits to one discrete decision: a direction to pursue, a backtrack to a previously visited waypoint, or a verified stop. The decision arrives wrapped in a structured JSON record carrying progress analysis and action reasoning.


\paragraph{System~1: fast policy.}
Once System~2 commits to a direction, the remaining challenge is to translate this high-level intent into fast perception-to-action execution. $\varphi_1$ is a lightweight grounding-specialized MLLM that observes the selected view along with the planner's target description and predicts a bounding box for the intended target, without relying on a pretrained waypoint predictor. This design preserves the ability to ground distant semantic cues, such as a hallway opening, which may not correspond to explicit intermediate waypoints. The deterministic controller $\pi_{\mathrm{robot}}$ then selects a representative pixel within the predicted box, back-projects it into 3D using the depth map and camera intrinsics, and transforms the resulting point into the world frame using the robot's odometry pose. Finally, it executes a short-horizon trajectory through fast-marching planning over a lightweight local traversability map.

\subsection{Hierarchical Agentic Memory}
\label{sec:method-memory}

The dual-process architecture separates deliberation from execution, but System~2 still requires navigation history beyond the current observation. \ourmethod\ therefore uses hierarchical agentic memory: working memory retains recent frames, while an instruction-progress record and a depth-grounded world graph maintain instruction progress, visitation history, semantic state, and failure evidence. At each decision step, \ourmethod\ retrieves a compact graph slice conditioned on the current subgoal.

\paragraph{Memory state.}
Formally, at decision waypoint $t$, let $W_t$ denote the frame stream retained from the most recent $K$ completed waypoint transitions. We use $L_t$ for the ordered instruction-progress record and $G_t$ for the long-term world graph accumulated during the episode. The current subgoal $q_t$ is derived from $L_t$, and the memory context supplied to System~2 is
\begin{equation}
M_t = W_t \oplus L_t \oplus \mathcal{R}(G_t,q_t),
\label{eq:memory-state}
\end{equation}
where $\mathcal{R}$ is the query-conditioned read operator defined in Section~\ref{sec:method-retrieval}, and $\oplus$ denotes composition into the planning context. The current panorama $O_t$ remains immediate perception and is supplied separately. The window size $K$ therefore defines a \emph{consolidation boundary}: visual history inside $W_t$ remains verbatim, whereas older observations can return only through a structured readout of $G_t$.

\paragraph{Sparse grounded world graph.}
We represent the persistent state as a typed place--object graph
\begin{equation}
G_t = \big(\mathcal{V}^{P}_t,\mathcal{V}^{O}_t,
\mathcal{E}^{PP}_t,\mathcal{E}^{PO}_t\big).
\label{eq:world-model}
\end{equation}
Place states are formed by incremental single-link clustering. Each place state in $\mathcal{V}^{P}_t$ aggregates spatial anchors, agent-authored semantic context, visitation statistics, and optional failure evidence. Each object state in $\mathcal{V}^{O}_t$ associates a depth-grounded category observation with open-vocabulary descriptions and soft attributes at the place/category level. Place--place edges $\mathcal{E}^{PP}_t$ encode navigational connectivity, while place--object edges $\mathcal{E}^{PO}_t$ encode semantic containment and its confidence. This representation allocates persistent state only to discovered places and entities while retaining the geometry and topology needed for recall and backtracking. 

\paragraph{Decision-coupled updates.}
The System~2 call produces a structured decision record $J_t=(a_t,w_t,r_t)$, where $a_t$ is the high-level action, $w_t$ contains place semantics, observed objects, and instruction-progress updates, and $r_t$ is the action rationale. Let $\Gamma(O_t,D_t,P_t)$ denote the grounded geometric evidence extracted from the current RGB-D observation and pose. The routine task-progress and world-model states evolve jointly as
\begin{equation}
\begin{aligned}
L_{t+1} &= \mathcal{U}_{L}(L_t,w_t),\\
G_{t+1} &= \mathcal{U}_{G}\!\left(G_t,w_t,\Gamma(O_t,D_t,P_t)\right),
\end{aligned}
\label{eq:memory-update}
\end{equation}
where $\mathcal{U}_{L}$ applies progress updates and $\mathcal{U}_{G}$ associates observations with places and objects, fuses repeated evidence, and updates topology. When $a_t$ is a backtrack, the event-specific update in Section~\ref{sec:method-backtrack} attaches $r_t$ as failure evidence. The agent authors $w_t$ and $r_t$, while grounding, association, projection, and graph maintenance are automatic operations. Grounding DINO~\citep{liu2023groundingdino} and SAM~\citep{kirillov2023sam} provide open-set detections and masks that are projected through depth and consolidated into place-level object states.

\paragraph{Functional views of hierarchical memory.}
\ourmethod\ exposes the state in Equation~\ref{eq:memory-state} through four functional views (Figure~\ref{fig:memory}):
\begin{itemize}
\item \textbf{Working memory} retains recent verbatim frames in $W_t$, while $O_t$ supplies immediate perception separately.
\item \textbf{Episodic memory} reads the visitation trajectory and temporal metadata in $G_t$, supplying recency to retrieval.
\item \textbf{Semantic memory} reads place--object content and topology in $G_t$, supplying relevance to retrieval.
\item \textbf{Reflection memory} returns failure evidence attached to retrieved places, conditioning later decisions on previously abandoned branches.
\end{itemize}
Salience in the retrieval score is an importance term based on landmark and object evidence; reflection is orthogonal to scoring and re-enters context only when its associated place is retrieved.

\subsection{Subgoal-Conditioned Memory Retrieval}
\label{sec:method-retrieval}

\paragraph{Subgoal-conditioned place ranking.}
At waypoint $t$, the first pending item in $L_t$ defines the current subgoal $q_t$. A local sentence encoder~\citep{xiao2023bge} embeds $q_t$ and the summary of each place state $p\in\mathcal{V}^{P}_t$. We score each place by
\begin{equation}
s_t(p) = \alpha\,\mathrm{rel}(q_t,p)
+ \beta\,\mathrm{rec}_t(p)
+ \gamma\,\mathrm{sal}(p),
\label{eq:score}
\end{equation}
where $\mathrm{rel}$ is query--place cosine similarity, $\mathrm{rec}_t(p)=\rho^{t-\tau(p)}$ measures recency from the last visit $\tau(p)$, and $\mathrm{sal}$ combines landmark agreement with object richness. We use $(\alpha,\beta,\gamma)=(1.0,0.3,0.3)$ and a per-waypoint decay $\rho=0.85$ in all experiments. These terms capture task relevance, temporal continuity, and perceptual importance, respectively.

\paragraph{Topology-aware memory readout.}
Using a seed budget $K_s=3$ and final place budget $K_p=5$, the read operator $\mathcal{R}(G_t,q_t)$ retains the $K_s$ highest-scoring places as seeds, expands them by one hop, and includes the current-place neighborhood before keeping at most $K_p$ places. The resulting place--object subgraph is serialized into $M_t$, including semantic content, relative geometry, topology, and any associated failure evidence.

\subsection{Grounded Failure Memory for Backtracking}
\label{sec:method-backtrack}

In partially observable navigation, an abandoned branch is not merely motion to undo; it is negative evidence about the compatibility between the current subgoal and an explored region. \ourmethod\ therefore models backtracking as a coupled control--memory transition. When System~2 returns to a previously visited waypoint $b_t$, the abandoned suffix $\mathcal{B}_t$ is mapped to its associated place states and the decision rationale $r_t$ is stored as grounded failure evidence. The return target remains waypoint-level to preserve executable geometry, while the evidence is place-level so that it can persist across repeated visits.

This evidence is advisory rather than prohibitive: it neither changes $s_t(p)$ nor blocks a place. It re-enters $M_t$ only when $\mathcal{R}(G_t,q_t)$ retrieves the associated region, allowing System~2 to reinterpret the earlier failure under the current subgoal. Backtracking therefore changes both the agent's pose and the memory state on which subsequent decisions are conditioned, closing the loop from action and failure to grounded write, conditional recall, and revised action.

Algorithm~\ref{alg:loop} summarizes how control, memory writing, retrieval, and backtracking interact over a navigation episode.

\begin{algorithm}[t]
\caption{One episode of \ourmethod}
\label{alg:loop}
\begin{algorithmic}[1]
\REQUIRE Navigation instruction $\mathcal{T}$
\STATE Initialize progress $L_0$ from $\mathcal{T}$; $G_0,W_0\gets\emptyset$
\FOR{each decision waypoint $t$}
  \STATE Receive the current observation $(O_t,D_t,P_t)$
  \STATE $q_t\gets\textsc{CurrentSubgoal}(L_t)$
  \STATE $M_t\gets W_t\oplus L_t\oplus \mathcal{R}(G_t,q_t)$
  \STATE $(a_t,w_t,r_t)\gets\varphi_2(\mathcal{T},O_t,M_t)$
  \STATE $L_{t+1}\gets\mathcal{U}_L(L_t,w_t)$
  \STATE $G_{t+1}\gets\mathcal{U}_G(G_t,w_t,\Gamma(O_t,D_t,P_t))$
  \IF{$a_t=\textsc{Stop}$}
    \STATE \textbf{return}
  \ELSIF{$a_t=\textsc{Backtrack}(b_t)$}
    \STATE $\mathcal{B}_t\gets$ trajectory suffix after $b_t$
    \STATE $\mathcal{F}_t\gets\textsc{Places}(\mathcal{B}_t)\setminus\{p(b_t)\}$
    \STATE $G_{t+1}\gets\textsc{MarkFailure}(G_{t+1},\mathcal{F}_t,r_t)$
    \STATE Return to the stored waypoint anchor $b_t$ via $\pi_{\mathrm{robot}}$
  \ELSE
    \STATE $g_t\gets\varphi_1(a_t,O_t)$
    \STATE Execute $g_t$ via $\pi_{\mathrm{robot}}(g_t,D_t,P_t,\mathbf{K}_{\mathrm{cam}})$
  \ENDIF
  \STATE $W_{t+1}\gets\textsc{UpdateWindow}(W_t,O_t;K)$
\ENDFOR
\end{algorithmic}
\end{algorithm}

\section{Experiments}
\label{sec:exp}

We first describe the evaluation protocol (Section~\ref{sec:exp-setup}) and compare \ourmethod\ against published trained and zero-shot methods on R2R-CE, RxR-CE, and HM3D-v2 ObjectNav (Section~\ref{sec:exp-main}). We then examine inference cost and raw-history window sensitivity (Section~\ref{sec:exp-cost}), ablate the memory components (Section~\ref{sec:exp-ablation}), and analyze backtracking through a real episode trace (Section~\ref{sec:exp-qual}).

\begin{table*}[t]
\centering
{\fontsize{9pt}{10pt}\selectfont
\setlength{\tabcolsep}{1.2pt}
\begin{tabular}{@{}lc|cccc|cccc@{}}
\toprule
\multirow{2}{*}{\textbf{Method}} & \multirow{2}{*}{\textbf{Source}} &
\multicolumn{4}{c|}{\textbf{VLN-CE R2R Val-Unseen}} &
\multicolumn{4}{c}{\textbf{VLN-CE RxR Val-Unseen}} \\
& & \textbf{NE} $\downarrow$ & \textbf{OSR}$\uparrow$ &
\textbf{SR} $\uparrow$ & \textbf{SPL}$\uparrow$ &
\textbf{NE} $\downarrow$ & \textbf{SR}$\uparrow$ &
\textbf{SPL} $\uparrow$ & \textbf{nDTW}$\uparrow$ \\
\midrule
\rowcolor[HTML]{E3F0FA}\multicolumn{10}{c}{\textbf{Supervised Learning (Training Method)}} \\
NaVid~\cite{zhang2024navid}                      & \textit{RSS'24} & 5.47 & 49.1 & 37.4 & 35.9 & 8.41 & 34.5 & 23.8 & --   \\
Uni-NaVid~\cite{zhang2024uninavid}               & \textit{RSS'25} & 5.58 & 53.3 & 47.0 & 42.7 & 6.24 & 48.7 & 40.9 & --   \\
NaVILA~\cite{cheng2024navila}                    & \textit{RSS'25} & 5.22 & 62.5 & 54.0 & 49.0 & 6.77 & 49.3 & 44.0 & 58.8 \\
StreamVLN~\cite{wei2025streamvln}                & \textit{ICRA'26} & 4.98 & 64.2 & 56.9 & 51.9 & 6.22 & 52.9 & 46.0 & 61.9 \\
JanusVLN~\cite{zeng2025janusvln}                 & \textit{ICLR'26} & 4.78 & 65.2 & 60.5 & 56.8 & 6.06 & 56.2 & 47.5 & 62.1 \\
OmniNav~\cite{xue2025omninav}                    & \textit{ICLR'26} & 3.74 & 74.6 & 69.5 & 66.1 & 3.77 & 73.6 & 62.0 & --   \\
InternVLA-N1~\cite{wei2025internvlan1}           & \textit{arXiv} & 4.83 & 63.3 & 58.2 & 54.0 & 5.91 & 53.5 & 46.1 & 65.3   \\
ABot-N0~\cite{chen2026abotn0}                    & \textit{arXiv} & 3.78 & 70.8 & 66.4 & 63.9 & 3.83 & 69.3 & 60.0 & -- \\
SPAN-Nav~\cite{liu2026spannav}                   & \textit{arXiv} & 4.07 & 75.3 & 66.3 & 59.3 & 4.20 & 69.7 & 60.1 & 67.9 \\
\midrule
\rowcolor[HTML]{E3F0FA}\multicolumn{10}{c}{\textbf{Zero-Shot (Training-Free)}} \\
InstructNav~\cite{long2024instructnav}          & \textit{CoRL'24} & 6.89 & 47.0 & 31.0 & 24.0 & --   & --   & --   & --   \\
Open-Nav~\cite{qiao2024opennav}                 & \textit{ICRA'25} & 6.70 & 23.0 & 19.0 & 16.1 & --   & --   & --   & --   \\
CA-Nav~\cite{chen2025constraint}                & \textit{TPAMI'25} & 7.58 & 48.0 & 25.3 & 10.8 & 10.4 & 19.0 & 6.0  & 13.5 \\
SmartWay~\cite{shi2025smartway}                 & \textit{IROS'25} & 7.01 & \underline{51.0} & 29.0 & 22.5 & --   & --   & --   & --   \\
GC-VLN~\cite{yin2025gcvln}                      & \textit{CoRL'25} & 7.30 & 41.8 & 33.6 & 16.3 & \underline{8.80} & \underline{33.8} & 13.8 & --   \\
LaViRA~\cite{ding2025lavira}                    & \textit{ICRA'26} & 6.54$\pm$0.27 & 48.7$\pm$2.1 & \underline{38.3}$\pm$0.6 & 28.3$\pm$0.9 & --   & --   & --   & --   \\
Three-Step Nav~\cite{zheng2026threestepnav}     & \textit{AISTATS'26} & \underline{5.87} & 39.0 & 34.0 & \underline{29.1} & 9.21 & 22.0 & \underline{16.1} & \underline{45.7} \\
HiMemVLN~\cite{lyu2026himemvln}                 & \textit{arXiv} & 6.65 & 36.0 & 30.0 & 26.9 & --   & --   & --   & --   \\
\textbf{\ourmethod\ }                      & \textit{Ours} &
\textbf{3.92}$\pm$0.15 & \textbf{78.7}$\pm$3.1 & \textbf{61.0}$\pm$1.7 & \textbf{48.1}$\pm$0.2 &
\textbf{6.27}$\pm$0.23 & \textbf{52.7}$\pm$1.5 & \textbf{35.7}$\pm$2.6 & \textbf{54.3}$\pm$0.3 \\
\bottomrule
\end{tabular}}
\caption{Main results on VLN-CE R2R and RxR val-unseen. Our numbers are mean~$\pm$~std over three seeds.}
\label{tab:vlnce}
\end{table*}

\subsection{Setup}
\label{sec:exp-setup}

\paragraph{Benchmarks and metrics.}

We evaluate \ourmethod\ on R2R-CE~\citep{krantz2020beyond}, RxR-CE~\citep{ku2020rxr}, and HM3D-v2 ObjectNav~\citep{ramakrishnan2021hm3d} using the same 100-episode subsets of the val-unseen splits as \citet{ding2026unilavira}. For these three benchmarks, \citet{ding2026unilavira} reported that the SR values reproduced on these subsets deviate from the corresponding full-split results by no more than 1.1 percentage points; the appendix summarizes these previously reported subset-to-full comparisons. For each benchmark, we conduct three independent runs on the corresponding subset and report the mean and standard deviation. For R2R-CE, we report navigation error (NE), oracle success rate (OSR), success rate (SR), and success weighted by path length (SPL)~\citep{anderson2018spl}. For RxR-CE, we report NE, SR, SPL, and normalized dynamic time warping (nDTW)~\citep{ilharco2019ndtw}. For HM3D-v2 ObjectNav, we report SR and SPL. In addition, we report the mean number of context tokens used per episode.




\paragraph{Models and configurations.}
The agent is training-free: the System~2 planner is \texttt{Gemini-3.1-Pro}, the System~1 grounding model is \texttt{Qwen3.6-35B}, and a fast-marching controller executes motion (Section~\ref{sec:method-interface}). The retrieval encoder is a local \texttt{bge-large-en-v1.5}~\citep{xiao2023bge} resident on each worker GPU, so per-step retrieval costs no API calls. A working-window size of $K{=}1$ is used consistently across all three benchmarks.

\subsection{Main Results}
\label{sec:exp-main}
\label{subsec:main_results}

Tables~\ref{tab:vlnce} and~\ref{tab:hm3d_eqa} compare \ourmethod\ with both trained and zero-shot methods reported in prior work.

\subsubsection{VLN-CE (R2R and RxR)}
\label{sec:exp-vlnce}

On R2R, \ourmethod\ outperforms the strongest previously reported training-free baseline across all metrics (Table~\ref{tab:vlnce}): NE decreases from $5.87$ to $3.92$\,m, whereas OSR, SR, and SPL increase by $27.7$, $22.7$, and $19.0$ percentage points, respectively. On RxR, the same configuration reduces NE from $8.80$ to $6.27$\,m and increases SR, SPL, and nDTW by $18.9$, $19.6$, and $8.6$ percentage points, respectively. These consistent gains demonstrate that the improvements extend to the RxR setting, which contains longer instructions, without benchmark-specific retuning. Although the resulting OSR of $78.7$ on R2R exceeds those of all supervised methods reported in the table, the strongest trained systems continue to achieve higher SR and SPL on both benchmarks because agentic exploration samples next directions under uncertainty rather than following a learned shortcut.

\subsubsection{HM3D-v2 ObjectNav}
\label{sec:exp-hm3d}

\ourmethod\ achieves an SR of $79.7\pm0.5$ and an SPL of $43.2\pm0.6$ on HM3D-v2 ObjectNav, as shown in Table~\ref{tab:hm3d_eqa}. Compared with the strongest previously reported training-free result for each metric, \ourmethod\ improves SR by $3.5$ points and SPL by $4.5$ points. These scores also exceed the best supervised results in the table by $2.7$ and $1.9$ points, respectively. The gains over both training-free and supervised methods suggest that agentic reasoning uses room-level semantic priors learned during MLLM pretraining to improve navigation success and path efficiency.


\begin{figure*}[t]
\centering
\includegraphics[width=1.0\textwidth]{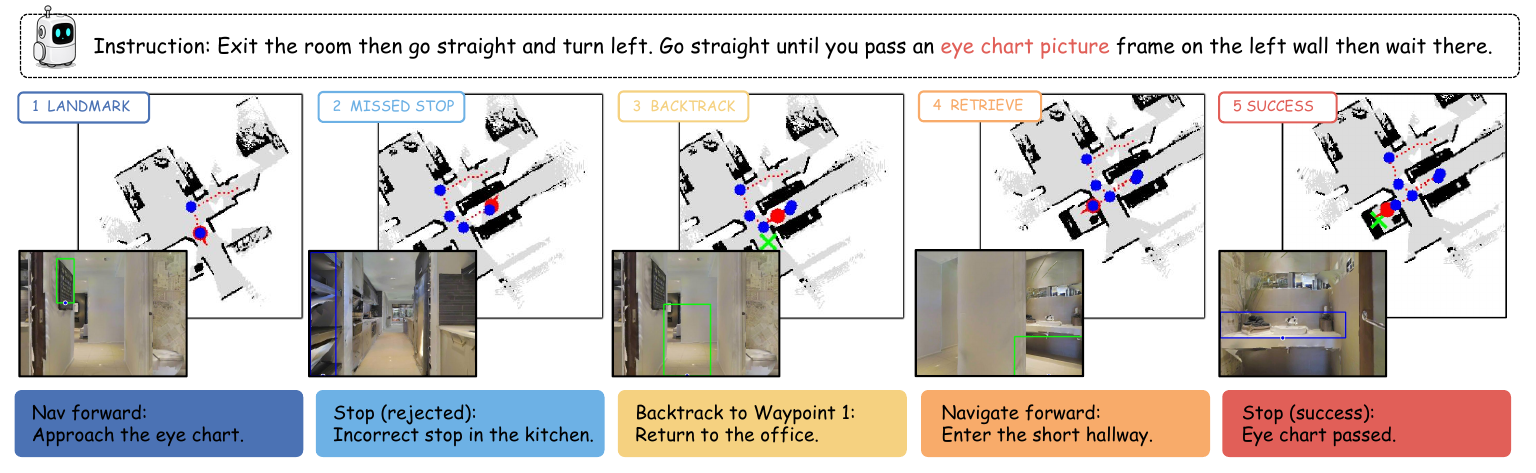}
\caption{An example of reflection memory for backtracking.}
\label{fig:episode-case}
\end{figure*}

\subsection{Inference Cost and Sensitivity to $K$}
\label{sec:exp-cost}
In Table~\ref{tab:context}, \ourmethod\ achieves $61.0$ SR and $48.1$ SPL, outperforming the best raw-history result for each metric by $2.7$ and $3.4$ percentage points, respectively. It uses $17.9$k System~2 tokens per decision and $244.9$k API tokens per episode. Compared with the lowest-cost raw-history setting ($K{=}3$), \ourmethod\ reduces these token counts by $45.1\%$ and $34.8\%$, respectively; compared with full history, the reductions are $69.0\%$ and $67.2\%$.

With raw history, expanding the window from $K{=}3$ to $K{=}5$, $K{=}10$, and full history monotonically increases both token counts without consistently improving SR or SPL. From $K{=}3$ to full history, API tokens per episode increase by $98.7\%$, while SR and SPL decrease by $4.7$ and $6.6$ percentage points, respectively. Larger visual contexts alone therefore do not replace task-relevant memory selection. Raw history retains overlapping, subgoal-irrelevant panoramas but lacks a place--object index for selective retrieval. In contrast, \ourmethod\ stores long-term navigation history in a depth-grounded world graph and retrieves a compact, subgoal-conditioned graph slice for each decision.

\begin{table}[!t]
\centering
{\fontsize{9pt}{10pt}\selectfont
\setlength{\tabcolsep}{2.5pt}
\begin{tabular*}{\columnwidth}{@{\extracolsep{\fill}}lcc@{}}
\toprule
\textbf{Method} & \textbf{SR}$\uparrow$ & \textbf{SPL}$\uparrow$ \\
\midrule
\rowcolor[HTML]{E3F0FA}\multicolumn{3}{c}{\textbf{Supervised Learning (Training Method)}} \\
DD-PPO~\cite{wijmans2020ddppo}           & 27.9 & 14.2 \\
Habitat-Web~\cite{ramrakhya2022habitatweb} & 57.6 & 23.8 \\
PIRLNav~\cite{ramrakhya2023pirlnav}      & 70.4 & 34.1 \\
OVRL-v2~\cite{yadav2023ovrlv2}           & 64.7 & 28.1 \\
Uni-NaVid~\cite{zhang2024uninavid}       & 73.7 & 37.1 \\
FiLM-Nav~\cite{yokoyama2025filmnav}      & 77.0 & 41.3 \\
\midrule
\rowcolor[HTML]{E3F0FA}\multicolumn{3}{c}{\textbf{Zero-Shot (Training-Free)}} \\
L3MVN~\cite{yu2023l3mvn}                 & 54.2 & 25.5 \\
VLFM~\cite{yokoyama2024vlfm}             & 62.6 & 31.0 \\
OpenFMNav~\cite{kuang2024openfmnav}      & 54.9 & 24.4 \\
InstructNav~\cite{long2024instructnav}   & 58.0 & 20.9 \\
SG-Nav~\cite{yin2024sgnav}               & 54.0 & 24.9 \\
ApexNav~\cite{kim2025apexnav}            & \underline{76.2} & 38.0 \\
DSCD-Nav~\cite{an2026dscdnav}            & 73.0 & \underline{38.7} \\
ReMemNav~\cite{wu2026rememnav}           & 67.8 & 36.6 \\
\textbf{\ourmethod\ (Ours)}              & \textbf{79.7}$\pm$0.5 & \textbf{43.2}$\pm$0.6 \\
\bottomrule
\end{tabular*}}
\caption{Main results on HM3D-v2 ObjectNav.}
\label{tab:hm3d_eqa}
\end{table}

\subsection{Ablations}
\label{sec:exp-ablation}

\begin{table}[t]
\centering
{\fontsize{9pt}{10pt}\selectfont
\setlength{\tabcolsep}{1.5pt}
\begin{tabular}{l|cc|cc}
\toprule
\textbf{Configuration} &
\textbf{SR}$\uparrow$ &
\textbf{SPL}$\uparrow$ &
\shortstack{\textbf{S2 tok.}\\\textbf{/ decision}$\downarrow$} &
\shortstack{\textbf{API tok.}\\\textbf{/ episode}$\downarrow$} \\
\midrule
\textbf{\ourmethod}
& \textbf{61.0$\pm$1.7} & \textbf{48.1$\pm$0.2} & \textbf{17.9k} & \textbf{244.9k} \\
\midrule
\noalign{\vskip 1pt}
\rowcolor[HTML]{E3F0FA}\multicolumn{5}{c}{\textbf{Raw-history window sensitivity}} \\
Raw history ($K{=}3$)
& 58.0$\pm$1.4 & 44.7$\pm$0.5 & 32.6k & 375.5k \\

Raw history ($K{=}5$)
& 54.3$\pm$2.1 & 38.4$\pm$0.3 & 39.8k & 457.8k \\

Raw history ($K{=}10$)
& 58.3$\pm$0.9 & 41.6$\pm$1.2& 48.7k & 534.0k \\

Raw history (Full)
& 53.3$\pm$1.2 & 38.1$\pm$0.7 & 57.7k & 746.1k \\
\bottomrule
\end{tabular}}
\caption{Controlled planner-memory comparison and inference cost on R2R-CE val-unseen.}
\label{tab:context}
\end{table}

Table~\ref{tab:ablation} ablates the world graph and its episodic, semantic, and reflection memory views. All variants use the same planner, grounding model, and controller. The two \emph{w/o WG} variants remove the graph and retain either the $K{=}1$ working window or the full raw visual history. The remaining variants keep the graph and remove EM, SM, RM, or all three; \emph{w/o RM} preserves backtracking but disables failure-note writing and serialization.

\paragraph{World graph.}
Removing the world graph causes the largest loss in spatial accuracy. With the $K{=}1$ working window, NE increases from $3.92$ to $4.97$\,m and SR decreases by $9.3$ points to $51.7$. Replacing the graph with full raw history increases SR to $53.3$, still $7.7$ points below the full model, and reduces SPL from $48.1$ to $38.1$.

\begin{table}[t]
\centering
{\fontsize{9pt}{10pt}\selectfont
\renewcommand{\arraystretch}{1.12}
\setlength{\tabcolsep}{2pt}
\begin{tabular*}{\columnwidth}{@{\extracolsep{\fill}}l|cccc@{}}
\toprule
\textbf{Configuration} & \textbf{NE}$\downarrow$ & \textbf{OSR}$\uparrow$ & \textbf{SR}$\uparrow$ & \textbf{SPL}$\uparrow$ \\
\midrule
\textbf{\ourmethod} & \textbf{3.92$\pm$0.15} & \textbf{78.7$\pm$3.1} & \textbf{61.0$\pm$1.7} & \textbf{48.1$\pm$0.2} \\
\midrule
\noalign{\vskip 1pt}
\rowcolor[HTML]{E3F0FA}\multicolumn{5}{c}{\textbf{World Graph}} \\
w/o WG ($K{=}1$) & 4.97$\pm$0.21 & 72.6$\pm$1.8 & 51.7$\pm$1.3 & 39.3$\pm$1.2 \\
w/o WG (Full) & 4.62$\pm$0.19 & 68.3$\pm$0.9 & 53.3$\pm$1.2 & 38.1$\pm$0.7 \\
\midrule
\noalign{\vskip 1pt}
\rowcolor[HTML]{E3F0FA}\multicolumn{5}{c}{\textbf{Agentic Memory}} \\
w/o EM & 4.53$\pm$0.37 & 72.0$\pm$3.6 & 53.3$\pm$2.6 & 41.6$\pm$1.4 \\
w/o SM & 4.72$\pm$0.29 & 74.0$\pm$2.7& 53.3$\pm$0.5 & 42.1$\pm$0.7 \\
w/o RM & 4.41$\pm$0.17 & 76.0$\pm$1.4 & 55.7$\pm$1.3 & 36.9$\pm$0.9 \\
w/o all & 4.63$\pm$0.22 & 75.7$\pm$1.9 & 50.7$\pm$1.7 & 31.5$\pm$1.1 \\
\bottomrule
\end{tabular*}}
\caption{Component ablations on R2R-CE val-unseen.}
\label{tab:ablation}
\end{table}

\paragraph{Agentic memory.}
Removing episodic or semantic memory reduces SR from $61.0$ to $53.3$, while SPL decreases to $41.6$ without EM and $42.1$ without SM. These results show that both temporal visitation cues and place--object relevance contribute to retrieval. Removing reflection memory reduces SR by $5.3$ points to $55.7$ and SPL by $11.2$ points to $36.9$, indicating that failure notes primarily improve path efficiency. Removing all three views reduces SR by $10.3$ points to $50.7$ and SPL by $16.6$ points to $31.5$. The graph alone is therefore insufficient; the planner also requires its episodic, semantic, and reflection memory views.

\subsection{Quantitative Analysis}
\label{sec:exp-qual}

\paragraph{Reflection memory.}
Across 100 R2R-CE episodes, the agent backtracked 175 times in 1{,}288 waypoint decisions, corresponding to 1.75 backtracks per episode and 13.6\% of decisions. Figure~\ref{fig:episode-case} shows the process. After a rejected stop in the kitchen, the agent records: ``I will backtrack to Waypoint 1 (office) to reorient myself and find the hallway with the eye chart picture frame on the left wall.'' At the hallway intersection, System~2 retrieves this note, selects the short hallway, passes the eye chart, and stops successfully.



\section{Conclusion}
\label{sec:conclusion}

We introduce \ourmethod, a framework that harnesses hierarchical agentic memory for zero-shot vision-and-language navigation. \ourmethod\ represents navigation history in a depth-grounded world graph and uses working, episodic, semantic, and reflection memory views to retain recent observations, retrieve spatial and semantic information, and reuse failure evidence. Without task-specific training, \ourmethod\ outperforms prior zero-shot methods and surpasses the full raw-history variant in navigation performance while requiring fewer inference-time tokens. Grounding agentic memory in a world graph allows \ourmethod\ to reuse episode-specific evidence across long trajectories without continually expanding raw visual history.

\bibliography{aaai2027}


\end{document}